\documentclass[fleqn,10pt,twocolumn]{AROB}

\usepackage{amsmath,amssymb,amsfonts}
\usepackage{algorithmic}
\usepackage{graphicx}
\usepackage{textcomp}
\usepackage{xcolor}

\usepackage[utf8]{inputenc}
\usepackage[doi=false,isbn=false,url=false,eprint=false,style=ieee,sorting=none]{biblatex} 
\title{Identifying Influential Actions in Human-Robot Interactions}

\author{Haoyang Jiang${}^{1,2}$, Chenfei Xu${}^{1,3}$, Yuya Okadome${}^{1,4}$, Yukata Nakamura${}^{1\dagger}$}
\speaker{Yukata Nakamura}

\affils{
${}^{1}$Guardian Robot Project, R-IH, Riken, Kyoto, Japan\\
${}^{2}$Faculty of Engineering, Monash University, Melbourne, Australia\\
${}^{3}$Graduate school of engineering science, Osaka University, Osaka, Japan\\
${}^{4}$Faculty of engineering, Tokyo University of Science, Tokyo, Japan\\
}

\abstract{%
Human-robot interaction combines robotics, cognitive science, and human factors to study collaborative systems. This paper introduces a method for identifying influential robot actions using transfer entropy, a statistic that measures directed information transfer between time series. TE is effective for capturing complex, nonlinear interactions. We apply this method to analyze how robot actions affect human behavior during a conversation with a remotely controlled robot avatar. By focusing on the impact of proximity, our approach demonstrates TE’s capability to identify key actions influencing human responses, highlighting its potential to improve the design and adaptability of robotic systems. }

\keywords{%
Human-robot Interaction, Transfer Entropy, Influential Actions, Interaction Dynamics
}

\begin{document}

\maketitle


\section{Introduction}\label{intro}
Human-robot interaction (HRI) is a multidisciplinary field that merges robotics, artificial intelligence, cognitive science, and human factors to study and develop systems where humans and robots can effectively work together. As robots increasingly become part of our daily lives, from service robots in healthcare and hospitality \cite{technologies9010008} to collaborative robots in industrial settings \cite{MUKHERJEE2022102231}, understanding the dynamics of these interactions is crucial. A key aspect of effective HRI is identifying and understanding the influential actions that drive these interactions, enabling the development of more intuitive and adaptive robotic systems.

One promising method for analyzing interactions between humans and robots is transfer entropy (TE) \cite{berger_transfer_2014} \cite{social_cue_jiang}. TE is a non-parametric statistic that measures the directed transfer of information between time series, capturing the influence of one variable on another over time. Unlike correlation-based methods, TE can detect asymmetrical and nonlinear relationships, making it particularly suitable for the complex and dynamic nature of HRI.

In this paper, we propose an approach to identify influential actions in HRI using transfer entropy. Our method involves capturing data from both human and robot agents during interaction, and then applying TE to quantify the influence of specific robot actions on human behavior. By focusing on the robot-to-human influence, we aim to uncover how robot actions guide human responses, providing a comprehensive understanding of the interaction dynamics from the robot's perspective.

\begin{figure}[!t]
\centerline{\includegraphics[width=0.6\linewidth]{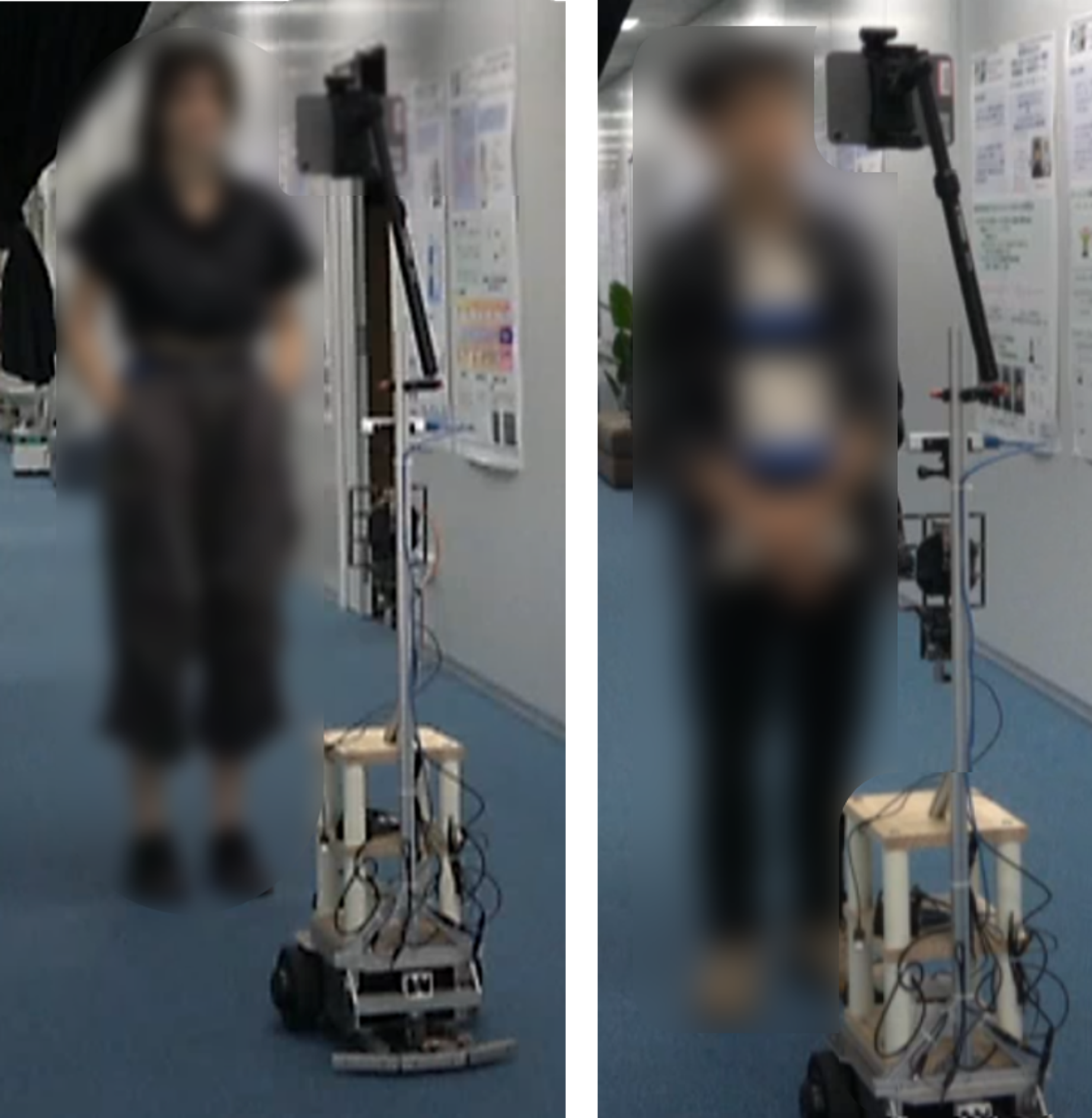}}
\caption{Human-Robot Avatar conversation.}
\label{fig:conversation}
\end{figure}

To validate our approach, we conducted an experiment involving a simple casual interaction between a human and a remotely controlled robot avatar (Figure \ref{fig:conversation}). During the interaction, the human and robot engaged in conversation while the robot performed basic movements, such as moving closer or farther and turning left or right. By analyzing the TE between the robot's actions and the human's responses, we focused on the influence of proximity on the interaction. Our findings demonstrate the effectiveness of using TE for finding influential actions in HRI, highlighting its potential to enhance the design and deployment of more responsive and adaptive robotic systems.

The remainder of this paper is structured as follows: Section 2 reviews related work in HRI and information-theoretic measures. Section 3 describes our experimental setup and data collection process. Section 4 presents our methodology for applying TE to HRI data. Section 5 discusses the results of our experiments and their implications. Finally, Section 6 concludes the paper and outlines future research directions.

\section{Background}\label{background}

\subsection{Human-Robot Interactions}
Human-Robot Interaction explores the dynamics between humans and robots, aiming to enhance the effectiveness of these interactions. Historically, HRI research has evolved from basic robot functionalities to more advanced collaborative and social interactions. Early studies focused on task execution, while recent research addresses complex interactions, such as collaborative robots and socially engaging systems \cite{Breazeal2016}. Key concepts include social influence, which examines how robot actions impact human behavior, and behavioral analysis, which studies how specific robot behaviors affect human responses \cite{DEGRAAF20131476} \cite{evaluating_Anzalone_2015}. Despite these advancements, identifying influential robot actions remains challenging due to the complexity of interactions \cite{mavrogiannis2023}. Recent developments in measurement techniques, behavioral modeling, and adaptive systems aim to address these challenges, offering improved methods to analyze and optimize robot actions in real-time \cite{maroto2023}. This study focuses on how robot actions influence human behavior, which is crucial for enhancing the design and deployment of responsive robotic systems.

\subsection{Transfer Entropy}
TE is a statistical measure used to analyze information transfer and potential causal relationships between two concurrent time series. In the field of economics, Baek et al. utilized TE to investigate the market influence of companies within the U.S. stock market \cite{baek_transfer_2005}. He and Shang examined various TE methods to study the relationships between nine stock indices from the U.S., Europe, and China \cite{he_comparison_2017}. TE has also been employed in the analysis of interactions, such as animal-animal or animal-robot interactions \cite{shaffer_transfer_2020}\cite{porfiri_inferring_2018}. Additionally, it has been used to study joint attention \cite{sumioka_causality_2007} and to model pedestrian evacuation dynamics \cite{xie_detecting_2022}. In the realm of robotics, Berger et al. applied TE to detect human-to-robot perturbations using low-cost sensors \cite{berger_transfer_2014}. Jiang et al. proposed a social cue detection and analysis framework using TE for general interactions, demonstrating its effectiveness in human-human interaction scenarios \cite{social_cue_jiang}. However, the framework has not yet been applied to human-robot interactions. In our study, we extend and adapt this framework to the context of human-robot interactions, aiming to validate its applicability and effectiveness in this domain.

\section{Experiments}\label{experiement}
\begin{figure}[!ht]
\centerline{\includegraphics[width=\linewidth]{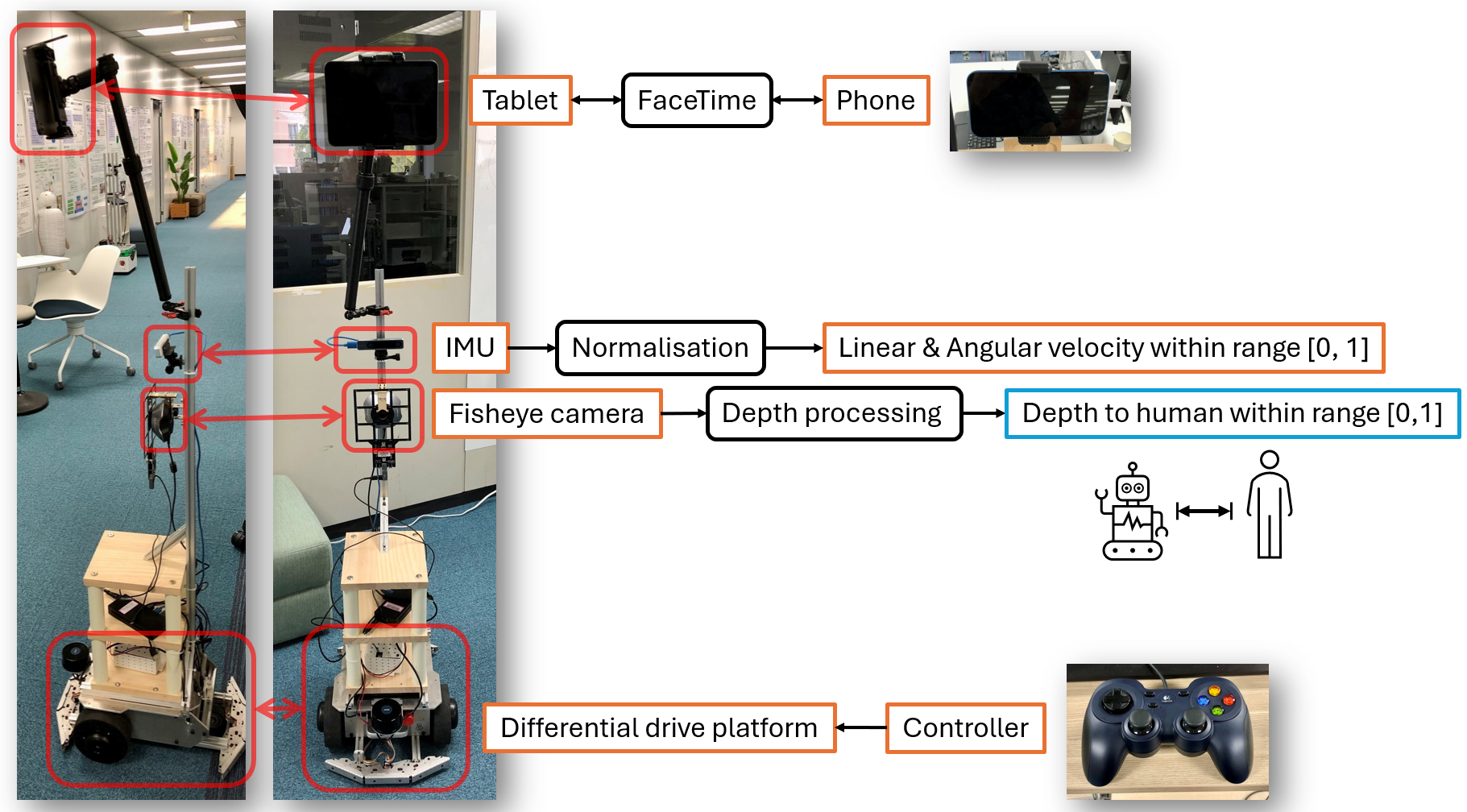}}
\caption{The robot avatar.}
\label{fig:robot_avatar}
\end{figure}

\begin{figure}[!ht]
\centerline{\includegraphics[width=\linewidth]{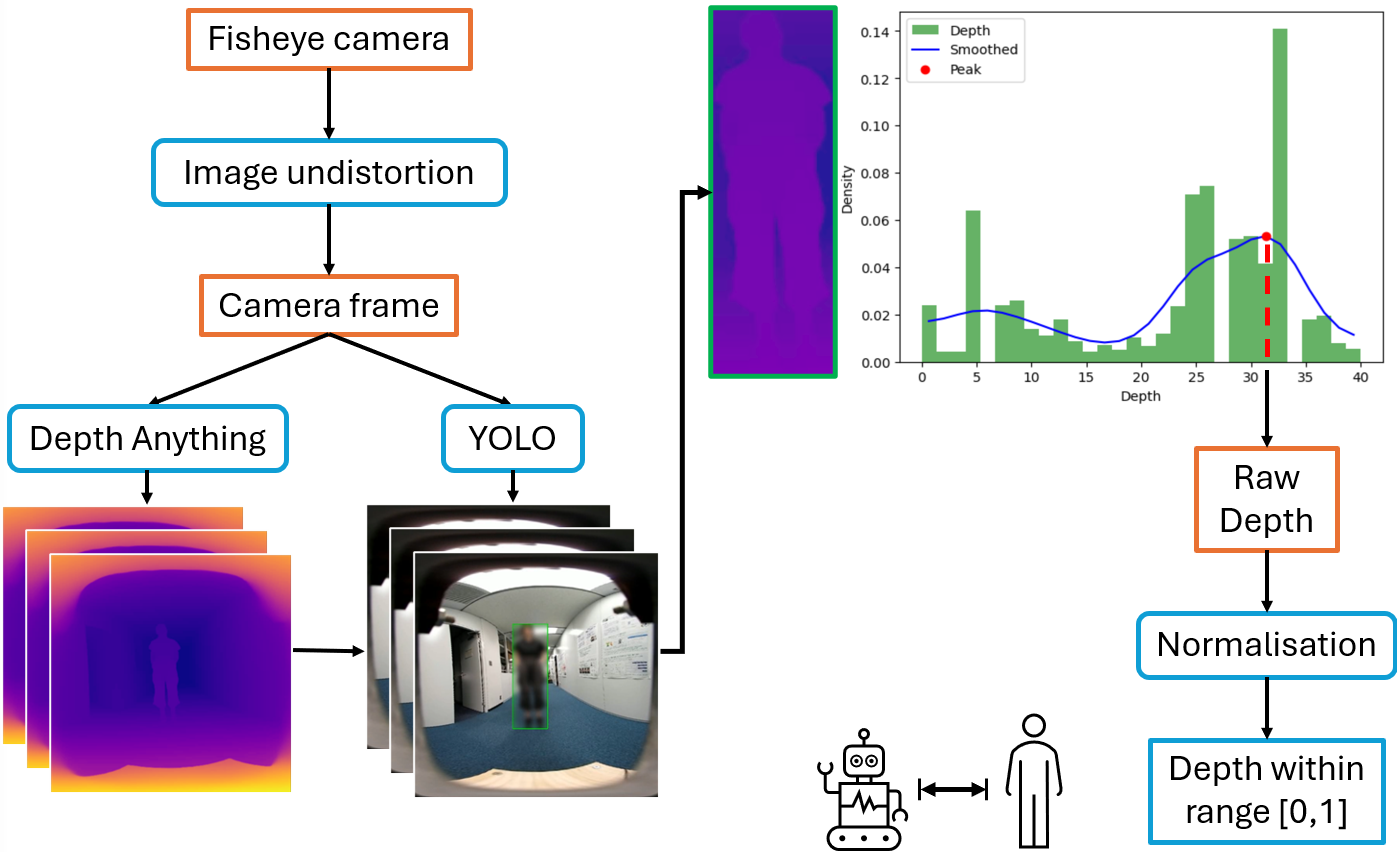}}
\caption{Depth processing.}
\label{fig:depth}
\end{figure}

In this study, we focus on a specific scenario where a remotely controlled robot avatar engages in conversation with a human participant to investigate the influence of the robot's actions on human behavior in terms of proximity. The robot avatar, depicted in Figure \ref{fig:robot_avatar}, is equipped with a tablet for video calling with the operator, a differential drive platform that provides three degrees of freedom and is remotely controlled by the operator using a controller. Additionally, the robot is outfitted with a Realsense T265 camera with an internal IMU sensor and a Xacti CX-MT500 fisheye camera. The IMU sensor supplies linear and angular velocity data, while the fisheye camera captures depth information relative to the human participant. The depth processing workflow, illustrated in Figure \ref{fig:depth}, utilizes YOLO \cite{glenn_jocher_2022_7347926} and Depth Anything V2 \cite{depth_anything_v2} to extract the bounded depth area of the human in the scene. We apply histogram and peak finding approaches to determine the relative depth between the robot and the human. Both the velocity data from the IMU sensor and the depth data are normalized between 0 and 1 before further analysis. This setup allows us to precisely measure and analyze the impact of the robot's proximity-related actions on the human participant.

Participants were invited to engage in face-to-face conversations with the operator via the robot avatar (Figure \ref{fig:conversation}). During these interactions, the operator performed four types of actions with the robot: moving forward (closer to the participant), moving backward (farther away from the participant), turning left in place, and turning right in place. Each action was executed multiple times throughout the conversation. The robot's actions were recorded using the IMU sensor, while the participant's relative depth to the robot was captured using the fisheye camera. To ensure natural interactions, participants were not given specific instructions prior to the experiment, other than to make themselves comfortable during the conversation. To minimize the influence of the conversation content on the participant's proximity, neutral topics were chosen, such as brief self-introductions, explanations of their research, and plans for the coming days. A total of six participants were recruited, resulting in approximately 15 minutes of data across eight experiments.

\section{Methodology}\label{methodology}
In this study, we adapt the framework proposed by Jiang et al. \cite{social_cue_jiang} with modifications to identify influential actions. This is achieved by measuring the influence of robot actions on human responses through information transfer using transfer entropy.

\subsection{Transfer Entropy}
Transfer entropy is fundamentally connected to the idea of Wiener-Granger causality, particularly under the conditions of linear Gaussian models, where they are equivalent \cite{granger_investigating_1969}. TE, denoted as $T_{Y\rightarrow X}$, represents the conditional mutual information between two variables, $X_t$ and $Y_t$ \cite{bossomaier_transfer_2016}. It can be mathematically expressed as follows.
\begin{equation}
    \label{eq:TE}
    \begin{aligned}
        T_{Y\rightarrow X}^{(k,l)}(t) & =I(X_t:{\mathbf{Y}}_{t}^{(l)}|{\mathbf{X}}_{t}^{(k)}) \\
        & =H(X_t|{\mathbf{X}}_{t}^{(k)})-H(X_t|{\mathbf{X}}_{t}^{(k)},{\mathbf{Y}}_{t}^{(l)})
    \end{aligned}
\end{equation}
In this context, $I$ represents mutual information, while $H$ stands for Shannon's entropy \cite{shannon_mathematical_1948}. The equation quantifies the information transfer from $Y$ to $X$, with $t$ denoting time, and $k$ and $l$ indicating the history lengths of $X_t$ and $Y_t$, respectively. In some studies, $X$ is referred to as the target, and $Y$ as the source. Intuitively, TE can be understood as the reduction in uncertainty about the state of $X$ when predicted solely from its own history, upon the introduction of an additional information source, $Y$ \cite{bossomaier_transfer_2016}. The central idea of the proposed framework is to utilize TE to gauge the transfer of information through observable robot actions to human behaviours.

\subsection{Influence of Actions}\label{sub:influential_actions}
\begin{figure}[!ht]
\centerline{\includegraphics[width=\linewidth]{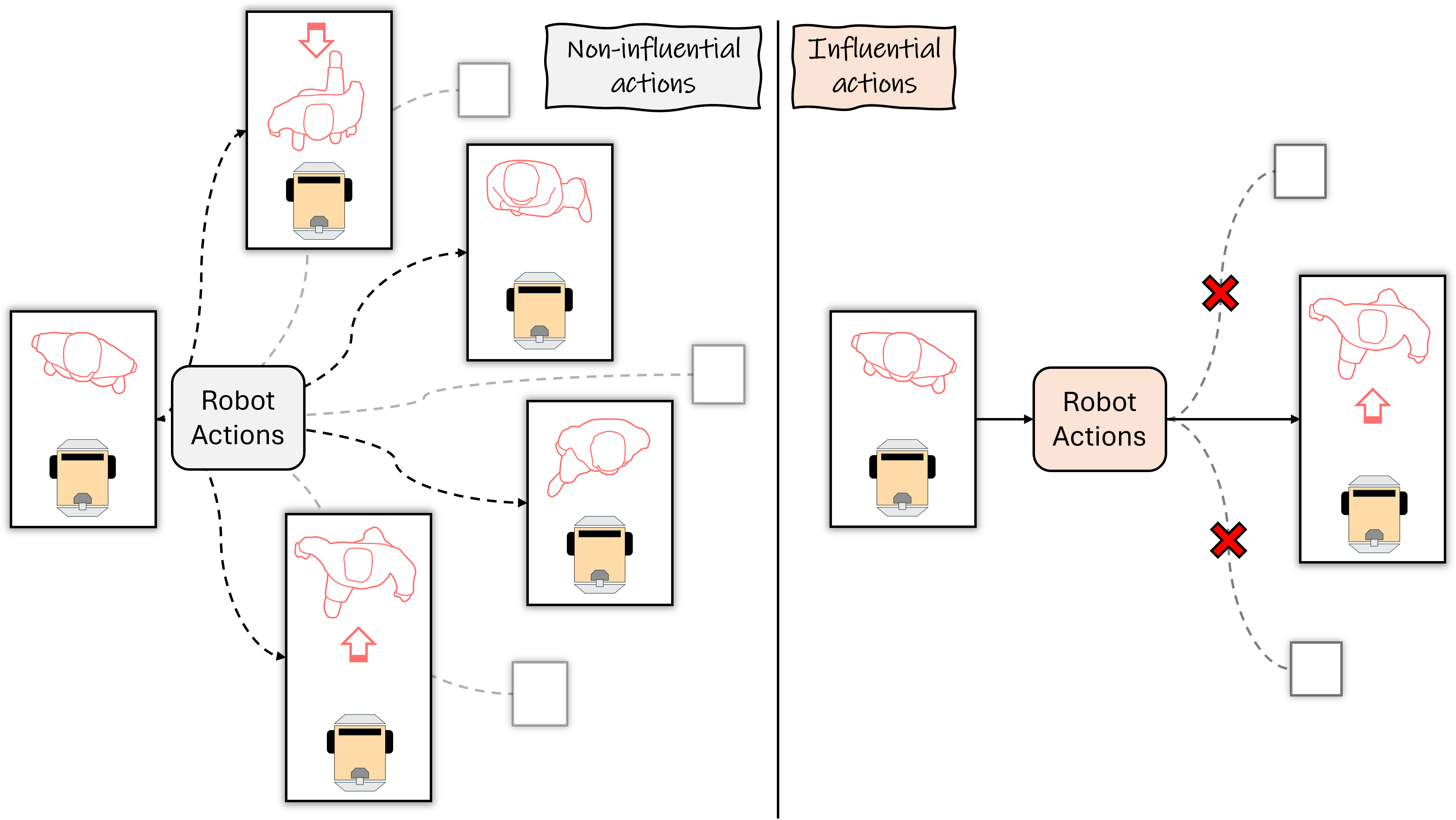}}
\caption{Left: Non-influential actions; Right: Influential actions.}
\label{fig:influence}
\end{figure}

The objective of this study is to identify influential actions of the robot in relation to the human participant, particularly in terms of proximity. Influential actions are defined as those that can reduce the uncertainty of future events. An illustration of influential actions is shown in Figure \ref{fig:influence}. By understanding these actions, we could enhance human-robot interaction, ensuring more predictable and effective collaboration.

\subsection{Workflow}
\begin{figure*}[!th]
\centerline{\includegraphics[trim={0.8cm 4.5cm 1.4cm 3.9cm},clip,width=0.7\linewidth]{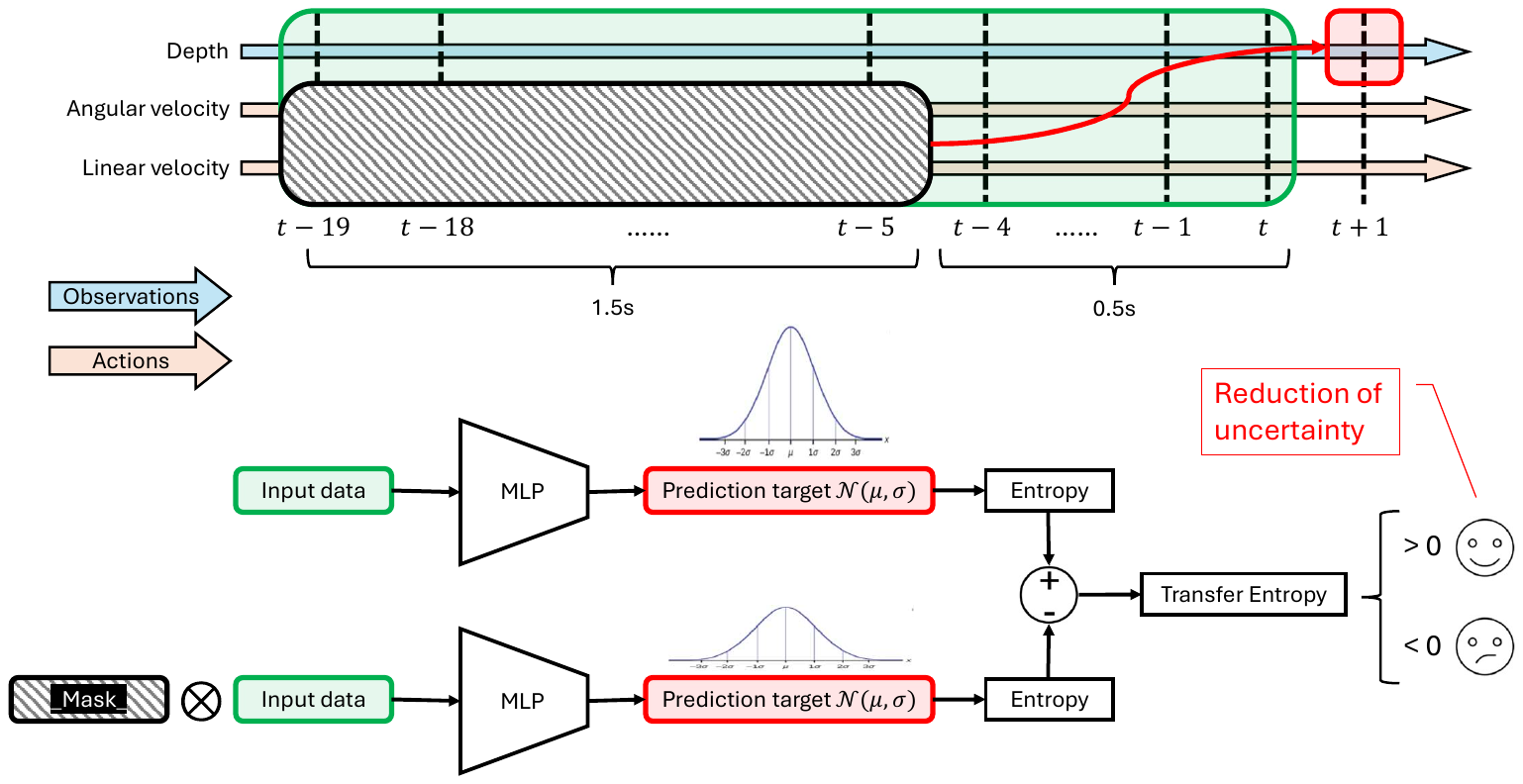}}
\caption{The workflow of the proposed method. Top: Visualisation of the timeline. Bottom: Illustration of the workflow.}
\label{fig:framework}
\end{figure*}

The workflow of the proposed method is illustrated in Figure \ref{fig:framework}. The robot's actions consist of its angular and linear velocity, while its observations are based on the relative depth measured by the camera. All data are time-series sampled at 10 fps. In this study, we model the future observation at $t+1$ using a fixed window of past observations and actions. The window size is two seconds, equivalent to 20 frames. The modeling method is flexible as long as the outcome is probabilistic. In our case, we selected a simple multilayer perceptron (MLP) with a Gaussian emission output layer, which produces a Gaussian random variable as the prediction. Denote the observation and action at $t$ as $o_t$ and $a_t$ respectively. The output with full window size can be expressed as a conditional Gaussian with mean $f({\mathbf{o}}_{t-19:t}, {\mathbf{a}}_{t-19:t})$ and covariance $f^{\mathbf{\sigma}}({\mathbf{o}}_{t-19:t}, {\mathbf{a}}_{t-19:t})$, 
\begin{equation}
    \label{eq:full_window}
    \begin{aligned}
        &P(o_{t+1}|{\mathbf{o}}_{t-19:t}, {\mathbf{a}}_{t-19:t}) \\
        &\sim {\mathcal{N}}_{o_{t+1}}(f({\mathbf{o}}_{t-19:t}, {\mathbf{a}}_{t-19:t}),f^{\mathbf{\sigma}}({\mathbf{o}}_{t-19:t}, {\mathbf{a}}_{t-19:t})).
    \end{aligned}
\end{equation}
We then calculate the uncertainty of the prediction as Shannon's differential entropy,
\begin{equation}
    \label{eq:diff_entropy}
    \begin{aligned}
        H({\mathbf{x}}) & =-\int p({\mathbf{x}})\log p({\mathbf{x}}) \,d{\mathbf{x}} \\
        & =\frac{D}{2}(1+\log (2\pi)) + \frac{1}{2}\log |{\mathbf{\sigma}}|.
    \end{aligned}
\end{equation}
In a parallel process, we repeat the same procedure using the same model employed for the full window size, but this time we apply a mask to zero out a specific period of past actions in the input. Specifically, we are interested in measuring the influence of past actions from $t-19$ to $t-5$ covering a duration of 1.5 seconds. Therefore, our mask covers the actions from $t-19$ to $t-5$. This masking results in a different prediction:
\begin{equation}
    \label{eq:marginal_window}
    \begin{aligned}
        &P(o_{t+1}|{\mathbf{o}}_{t-19:t}, {\mathbf{a}}_{t-4:t}) \\
        &\sim {\mathcal{N}}_{o_{t+1}}(f({\mathbf{o}}_{t-19:t}, {\mathbf{a}}_{t-4:t}),f^{\mathbf{\sigma}}({\mathbf{o}}_{t-19:t}, {\mathbf{a}}_{t-4:t})).
    \end{aligned}
\end{equation}
To compute TE, we calculate the difference between the two entropies,
\begin{equation}
    \label{transfer_entropy}
    \begin{aligned}
        &T_{{\mathbf{a}}_{t-19:t-5}\rightarrow o_{t+1}}(t) \\
        & =I(o_{t+1}:{\mathbf{a}}_{t-19:t-5}|{\mathbf{o}}_{t-19:t}, {\mathbf{a}}_{t-4:t}) \\
        & =H(o_{t+1}|{\mathbf{o}}_{t-19:t}, {\mathbf{a}}_{t-4:t})-H(o_{t+1}|{\mathbf{o}}_{t-19:t}, {\mathbf{a}}_{t-19:t}).
    \end{aligned}
\end{equation}
This TE represents the change in uncertainty of the future outcome at $t+1$ when we have knowledge of the masked period of actions from $t-19$ to $t-5$, compared to the scenario where this knowledge is absent. A positive TE value indicates that the masked period of actions helps reduce the prediction's uncertainty, signifying that these actions are influential by our definition in Section \ref{sub:influential_actions}.

\begin{figure*}[!ht]
    \centering
    \includegraphics[width=0.8\linewidth]{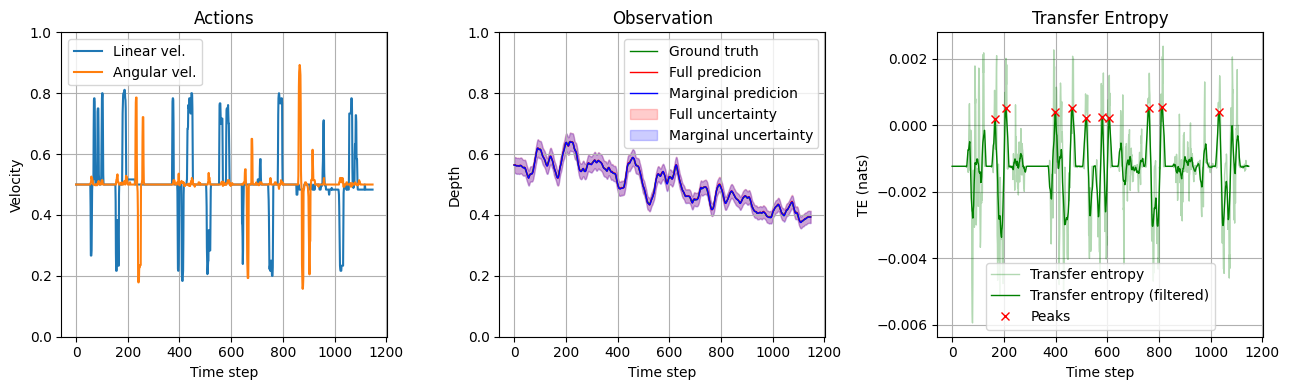}
    \caption{TE measurements for one experiment. Left: Action signals, including linear and angular velocity. Center: Depth measurements with predictions from both processes. The full prediction is based on the complete window of actions, while the marginal prediction uses masked past actions. Both the mean and uncertainty ($\pm3$ standard deviations) of the predictions are shown. Right: Transfer entropy measurements with highlighted peaks.}
    \label{fig:TE-result}
\end{figure*}

\subsection{Model Structure and Training}
Our model is a Multi-Layer Perceptron (MLP) with a Gaussian emission output layer. The input size is 80, followed by two hidden layers with sizes of 32 and 16, respectively. The output layer has a size of 2, representing the mean and standard deviation of the Gaussian distribution. During training, we randomly apply the mask at each epoch, allowing us to use a single model for both processes. This approach ensures that the model learns to adapt to varying levels of information availability. We use Negative Log-Likelihood (NLL) as the loss function for training.

\subsection{Compensation for Relative Measurements}
The depth measured using the robot's camera represents a relative measurement between the robot and the human, but it does not capture true changes in the global frame. For instance, a decrease in the measured depth could result from either the human moving towards the robot or the robot moving towards the human. To accurately assess the influence of the robot's actions on the human participant, we need to focus on the true changes in the global frame rather than just relative measurements. Therefore, we incorporate a segment of past actions ${\mathbf{a}}_{t-4:t}$, specifically 0.5 seconds, into the input for both processes. This inclusion helps compensate for the relative nature of the depth measurements by providing context about the genuine global changes, thus converting the relative observations into global observations.

\section{Results}\label{Disucssion}
We apply the proposed framework to each experiment to calculate the TE value at each time step and then use a low-pass filter to smooth the TE signal. For TE values that are positive, we identify the local maxima. Figure \ref{fig:TE-result} displays the transfer entropy measurements from one experiment, where each TE local maximum at $t+1$, referred to as a TE peak, corresponds to a sequence of influential actions from $t-19$ to $t-5$.

\begin{figure}[!th]
    \centering
    \includegraphics[width=1\linewidth]{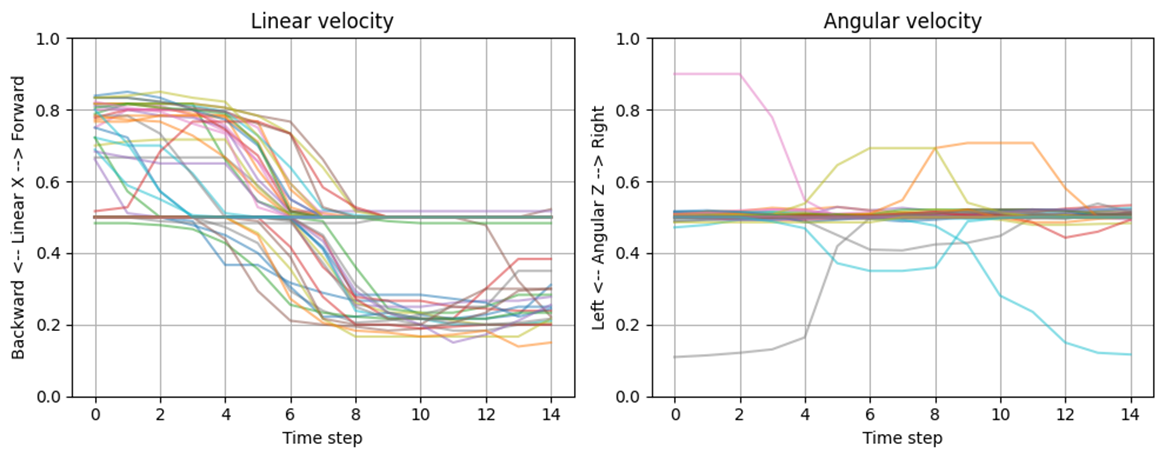}
    \caption{All identified influential action sequences.}
    \label{fig:all-actions}
\end{figure}

\begin{figure}[!th]
    \centering
    \includegraphics[width=1\linewidth]{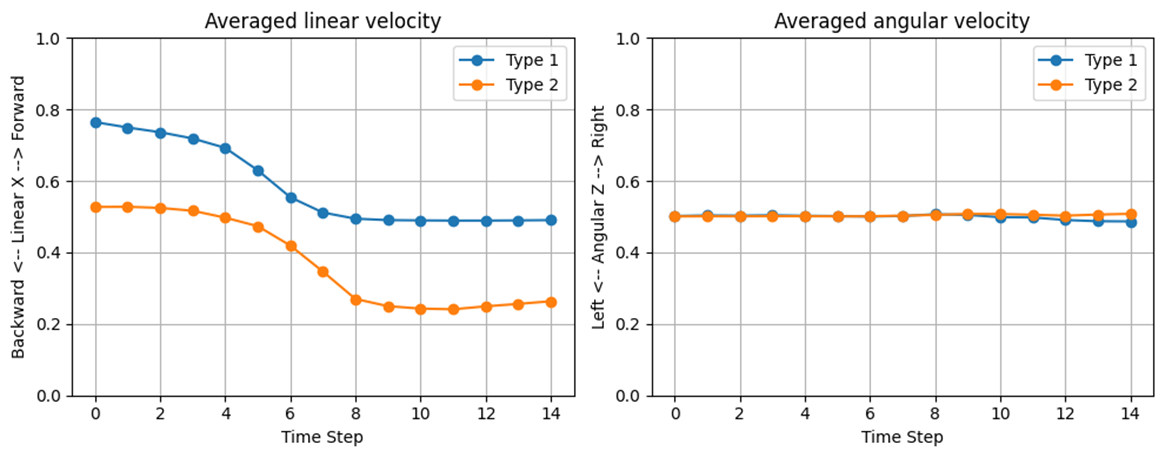}
    \caption{Averaged influential action sequences.}
    \label{fig:key-actions}
\end{figure}

Figure \ref{fig:all-actions} shows the influential action sequences identified from TE peaks across all experiments. The action signals are normalized between 0 and 1, where 0.5 indicates zero velocity. For linear velocity, values above 0.5 denote forward movement, while values below 0.5 indicate backward movement. For angular velocity, values above 0.5 represent anticlockwise rotation, and values below 0.5 indicate clockwise rotation. We observe that most angular velocity values remain near zero (0.5), suggesting that rotational actions are not significant in influencing proximity. However, we identify two distinct types of linear velocity sequences. We use K-means clustering \cite{kmeans1982} with Dynamic Time Warping (DTW) for distance calculation to group the linear velocity actions into two groups. By averaging these two groups of linear velocity sequences, Figure \ref{fig:key-actions} provides a clearer picture of these typical patterns. Type 1 refers to the end of a moving forward action, and type 2 refers to the start of a moving backward action. 

\begin{figure*}[!th]
    \centering
    \includegraphics[width=1\linewidth]{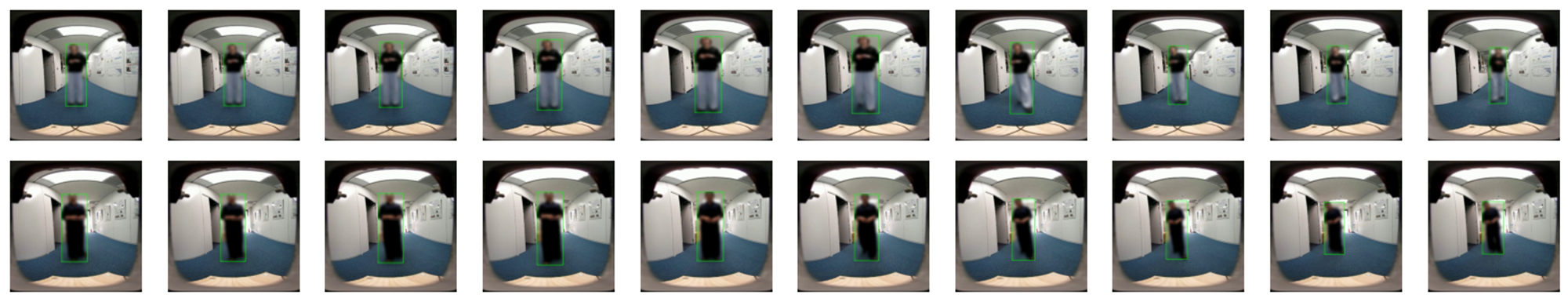}
    \caption{Two example sequences of Type 1 actions: The robot moves forward while the participants move backward.}
    \label{fig:forward}
\end{figure*}

\begin{figure*}[!th]
    \centering
    \includegraphics[width=1\linewidth]{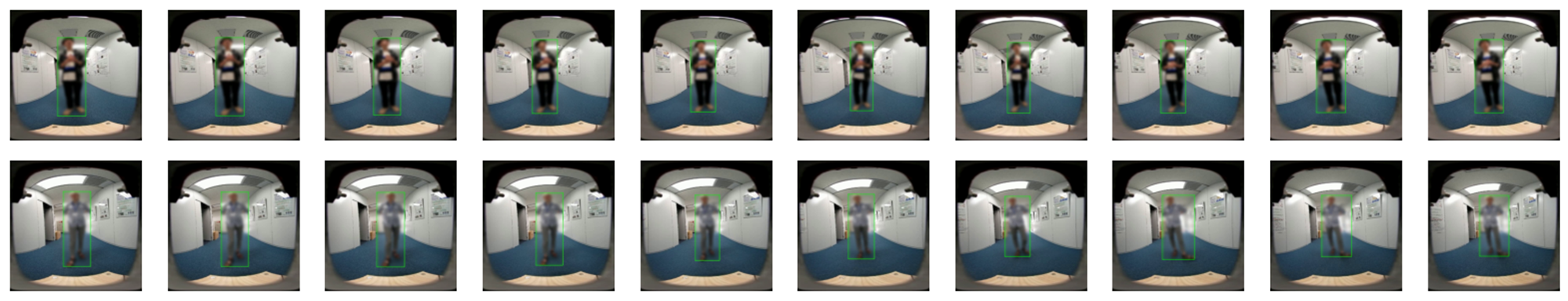}
    \caption{Two example sequences of Type 2 actions: The robot moves backward while the participants catch up.}
    \label{fig:backward}
\end{figure*}

Type 1 actions influence the human participant by encroaching upon their personal or intimate space, while Type 2 actions affect the participant by exiting their social space during a conversation \cite{hall_hidden_1966}. Examples of these two action types are illustrated in Figures \ref{fig:forward} and \ref{fig:backward}. Notice that in the case of Type 1 actions, where the robot moves forward and the participant adjusts by moving backward, participants typically react after the robot’s forward movement has ceased. Conversely, in Type 2 actions, where the robot moves backward and the participant advances to close the distance, participants tend to respond immediately as the robot begins its backward movement. This temporal distinction is also reflected in the framework, which captures Type 1 actions as marking the end of the forward movement and Type 2 actions as indicating the beginning of the backward movement.

\section{Conclusion}\label{conclusion}
In this work, we introduce a method for identifying influential robot actions in a human-robot avatar conversation, specifically focusing on the aspect of proximity. We demonstrate that the proposed framework effectively identifies typical robot actions that impact human proximity and successfully captures the temporal characteristics of these interactions. Additionally, this work extends and adapts the existing framework \cite{social_cue_jiang} from human-human interactions to human-robot interactions, providing a more comprehensive understanding of how robot behaviors influence human responses in authentic settings.

A considerable amount of future work remains to be explored. First, the dataset used in this study is relatively small. Expanding it to include a larger and more diverse range of interactions could enhance the robustness and generalizability of the findings. Additionally, investigating different contexts and scenarios may provide deeper insights into human-robot interaction dynamics and improve the framework's applicability across various settings. Second, exploring alternative modeling strategies that are better suited for temporal causal reasoning, such as Vector Autoregression \cite{var1980} and Causal Bayesian Networks \cite{pearl2009}, could be beneficial. Third, the current robot design is quite basic as a robot avatar. Enhancing its appearance could facilitate more natural interactions. Incorporating features like independent head movement, a crucial element in human-human interactions, would be particularly interesting \cite{kleinsmith2013}. Further, collecting and analyzing audio signals during conversations could offer insights into the influence of auditory cues. Lastly, leveraging influence measurements as metrics could potentially guide the robot's decision-making process, providing greater control over interaction dynamics.

\AtNextBibliography{\small}
\printbibliography

\end{document}